# An EnKF-LSTM Assimilation Algorithm for Crop Growth Model

Siqi Zhou, Ling Wang, Jie Liu, *Fellow, IEEE,* Jinshan Tang, *Senior Member, IEEE*

*Abstract*—Accurate and timely prediction of crop growth is of great significance to ensure crop yields and researchers have developed several crop models for the prediction of crop growth. However, there are large difference between the simulation results obtained by the crop models and the actual results, thus in this paper, we proposed to combine the simulation results with the collected crop data for data assimilation so that the accuracy of prediction will be improved. In this paper, an EnKF-LSTM data assimilation method for various crops is proposed by combining ensemble Kalman filter and LSTM neural network, which effectively avoids the overfitting problem of existing data assimilation methods and eliminates the uncertainty of the measured data. The verification of the proposed EnKF-LSTM method and the comparison of the proposed method with other data assimilation methods were performed using datasets collected by sensor equipment deployed on a farm.

## I. INTRODUCTION

DUE to the development of the world economy and the continuous growth of the population, the global demand for food is expected to continue to rise in the future. However, there remain hurdles to be addressed and surmounted, including the ability to produce fertilizers, the availability of land and water resources, and the consequences of global climate change. Consequently, accurate and timely prediction of crop growth is therefore of great significance for ensuring crop yields.

There have been significant advances in modelling crop growth and development using mechanistic models. These models simulate photosynthesis, gas exchanges among the canopy and the atmosphere, phenology, soil moisture and temperature dynamics, biomass growth and grain yield formation to predict the evolution of the crop from sowing to harvest. There is usually a discrepancy between the simulation results of crop models and the observed results because the prospective simulation of crop models can lead to an accumulating discrepancy.

Data assimilation methods have been extensively studied for crop growth models, including the WOFOST model. However, the accuracy of assimilation results can be significantly compromised by the quality of observed data. During field data collection, several factors such as environmental conditions, equipment limitations, and inconsistent collection standards can affect the quality of the data. Moreover, there are occasions when observation data is not available on certain days. As a result, there is a need for assimilation methods that can effectively handle low-quality observation data to enhance the accuracy of crop growth models.

Therefore, we propose a data assimilation method that combines the LSTM (Long Short-Term Memory) model with the Ensemble Kalman filter to tackle the challenges associated with low-quality observation data. This integrated approach will notably enhance the model's performance by generating more precise estimates of the state variables. Consequently, the model will offer improved output for crop growth estimation and yield prediction.

The innovations of this paper are as follows:

- This paper introduces a novel data assimilation algorithm for crop growth models, which addresses the challenges of missing and incorrect observation data. The proposed algorithm integrates the Ensemble Kalman Filter (EnKF) and Long Short-Term Memory (LSTM) model to estimate the state variables of the WOFOST crop growth model.
- This approach can be applied to other crop growth models as well. Moreover, the proposed crop growth model is effective for many different crops, successfully predicting the growth of various crop varieties including rice, maize, and soybeans.
- The proposed method is validated using data collected from sensor devices deployed in a farm. The data was gathered throughout a complete crop growth cycle in the most recent years, encompassing three crop types, namely rice, maize, and soybeans.

The paper is organized as follows: Section 2 discusses the work related to research. Section 3 discusses the structure of the crop growth model and DA method studied in this paper. Section 4 introduces the experimental content, and section 5 discusses the future development and draws a conclusion.

Siqi Zhou is with the Department of Computer Science and Technology, Harbin Institution of Technology, Heilongjiang, China. (e-mail: zhousiqi0530@qq.com).
Ling Wang is with the Department of Computer Science and Technology, Harbin Institution of Technology, Heilongjiang, China. (e-mail: wangling@hit.edu.cn).
Jie Liu is with the Department of Computer Science and Technology, Harbin Institution of Technology, Heilongjiang, China. (e-mail: jieliu@hit.edu.cn).
Jinshan Tang is with George Mason University, Fairfax, VA, 22033, USA. (e-mail: jtang25@gmu.edu).

## II. RELATED WORKS

Data assimilation was first proposed by Charney. Since then, it was gradually applied to atmospheric circulation models, such as numerical weather prediction, ocean circulation models, and land surface models. In DA, either the state of the system (e.g., the parameters that describe the system at a given point and location), the model parameters (often assumed to be constant throughout the time of the model run), or the initial conditions of some processes are assumed to be random variables, defined with a probability density function (pdf). The shape of the pdf critically encodes the uncertainty in our belief in the value of the parameters or state. DA methods provide a way to phrase the combination of observations (e.g., evidence) and models in an optimal way, by generally using Bayes' rule to update a prior pdf (e.g., predictions from a crop growth model) when evidence (e.g., uncertain EO data-derived parameters) is available. A few different methods have been developed to do this Bayesian update; their relative merits are usually based on the assumptions made to solve for the a posteriori (analysis) parameter/state pdf.

Traditional data assimilation methods widely used in agriculture can be divided into two categories: continuous data assimilation algorithms and sequential data assimilation algorithms. The continuous data assimilation algorithm takes the 4DVar algorithm as an example, which needs to define an assimilation time window and uses all the observed data and simulated state values in the assimilation window to make an optimal estimation. The initial field of the model is adjusted continuously through iteration, and the model trajectory is finally fitted to all the observed values obtained in the assimilation window period. When the difference between the practical value and the model's predicted value is slight, the difference between the corrected trajectory and the predicted trajectory of the initial model is slight. However, when the difference between the observed value and the model's predicted value is significant, the difference between the corrected trajectory and the predicted trajectory of the initial model is also significant.

Sequential data assimilation algorithm takes Kalman filter algorithm as an example, including prediction and update. In the prediction phase, the model is initialized according to the state value at time t and integrated forward until new observations are entered. The updating process is to weight the observed value at the time t+1 and the predicted value of the model state to obtain the optimal state estimate at the time and re-initialize it. Repeat the two steps until you have completed state predictions and updates for all observed data moments. Based on the traditional Kalman filtering algorithm, Kalman filtering has been widely developed and improved. For example, the Ensemble Kalman filter algorithm, which simulates the original covariance matrix with sample covariance, replaces the original covariance matrix with a small multi-magnitude matrix operation to update the original covariance matrix, which has a good effect when applying the model of higher magnitude that is not applicable to traditional Kalman filter.

For agricultural data, EnKF can generate initial conditions for integrated prediction and is easier to implement than the 4DVar algorithm. The above DA methods either have a direct analytic solution (e.g. the Kalman filter and derivatives) or can be solved by minimizing a cost function(4DVar). In either case, the use of Gaussian distributions results in a solution that can be encoded as a mean vector and a covariance matrix. In cases involving non-linear observation operators and crop growth models, or when uncertainties in either the model or observations are non-Gaussian, the assumption of normality in the posterior could be inadequate. In these cases, approaches based on sampling are preferred.

Compared with the widely used EnKF, Markov chain Monte Carlo (MCMC) allows the propagation of non-Gaussian distributions through complicated crop growth models by particle filters. MCMC methods are practical for low dimensional problems. If the dimensionality is high, MCMC methods are slow to explore the solution space, and convergence is hard to achieve in practical time scales. Thus, neither EnKF nor MCMC is effective when dealing with non-Gaussian distributions. We propose to use neural networks, which have strong nonlinear system approximation abilities and high-dimensional feature extraction capabilities. Neural networks have been successfully applied to data assimilation in other fields, such as wind-wave prediction and nuclear energy prediction. In our research, we propose a new combination of traditional DA method and neural networks to tackle the problem of non-Gaussian distributions through crop growth models within practical converge timing. We developed a lightweight data assimilation method which combines EnKF with LSTM for improving crop growth prediction. Our approach employs a recurrent neural network (RNN) to learn the data assimilation process of the EnKF method and addresses nonlinear issues typically encountered in traditional methods of agricultural data processing. With our proposed method, accurate assimilation of predicted data converges at less timing scale for high dimensional data. We demonstrate the effectiveness of our method by applying it to the WOFOST model for the prediction of crop growth.

## III. METHODS

### A. WOFOST Model

The WOFOST model (World FOod STudies) is a simulation model designed for the quantitative analysis of annual field crop growth and production. Yield under nutrient restriction is calculated based on soil characteristics and water restriction conditions. The model can simulate three different yield levels: potential yield limited by light temperature, rain-fed yield limited by light warm water, and yield limited by light warm water fertilizer. It is driven by daily meteorological data to limit and adjust the growing process of crops through soil, management, and crop parameter data.

LAI (Leaf Area Index), which means the multiple of the total area of plant leaves in the land area per unit of land area, represents the crop's ability to capture light and assimilate carbon, which are crucial indicators of the potential grain yield and thus the prediction of LAI is very important. In this paper, we study the application of WOFOST to predict the LAIs of crops. For the prediction of the LAI value at time (t+ Δt) using WOFOST model, the inputs of the model include two types of data: one is the LAI value at the time t, the other type of inputs are the soil data, crop data, agromanagement data, and weather data. The specific measurements for soil, crop, agromanagement, and weather are listed in Table 1. For example, the specific measurements for weather include the maximum temperature of the day, the minimum temperature of the day, daily total radiation, vapour pressure, wind speed, and precipitation. Let $P(t)$ be a vector whose elements are the measurements of soil, crop, agromanagement, and weather at time t. The prediction of LAI at time t+1 from the predicted LAI value at time earlier than t+1 can be done by

$$L(t+1) = WOFOST(L(0), L(1), ..., L(t), P(t)) \quad (1)$$

where L (t+1) is the predicted LAI value at time (t+1). See Figure 1 for the structure. When t=0, L(0) =0, which is the initialized value.

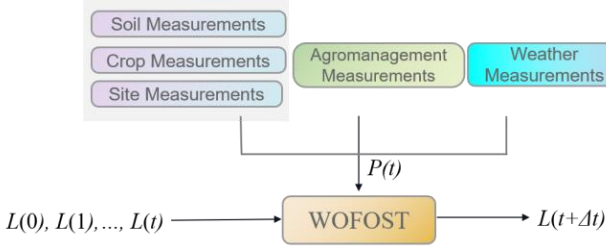

**Fig. 1.** WOFOST model for LAI prediction.

TABLE I
DIFFERENT MEASUREMENTS USED FOR LAI PREDICTION

| Name of Measurements | Specific Measurement | Description | Unit |
|---|---|---|---|
| Soil measurements | SMTAB | soil moisture content | pF |
| | CONTAB | 10-log hydraulic conductivity | pF |
| | SOPE | maximum percolation rate root zone | cm/day-1 |
| | KSUB | maximum percolation rate subsoil | cm/day-1 |
| | SPADS | 1st topsoil seepage parameter deep seedbed | / |
| | SPODS | 2nd topsoil seepage parameter deep seedbed | / |
| | SPASS | 1st topsoil seepage parameter shallow seedbed | / |
| | SPOSS | 2nd topsoil seepage parameter shallow seedbed | / |
| Crop measurements | TBASEM | lower threshold temp. for emergence | cel |
| | TSUM1 | temp. sum from emergence to init. beet growth | cel·d |
| | TDWI | initial total crop dry weight | kg/ha |
| | RGRLAI | maximum relative increase in LAI | ha/ha·d |
| | SPAN | life span of leaves growing at 35 Celsius | d |
| | TBASE | lower threshold temp. for ageing of leaves | cel |
| Site measurements | WAV | Initial soil water content | cm3/cm3% |
| | CO2 | Carbon dioxide concentration in the atmosphere | ppm |
| | SMLIM | Initial maximum moisture content in initial rooting depth zone [0-1] | / |
| Agromanagment measurements | Crop name | The name of the simulated crop | / |
| | Variety name | The variety of the simulated crop | / |
| | Crop start date | The start data of the simulation process | / |
| | Crop end date | The end data of the simulation process | / |
| | Max duration | The maximum duration of the simulation process | d |
| Weather measurements | TMax | Maximum temperature of the day | ℃ |
| | TMin | Minimum temperature of the day | ℃ |
| | IRRAD | Daily total radiation | J/m2/day |
| | VAP | vapor pressure | hPa |
| | WIND | wind speed | M/sec |
| | RAIN | precipitation | cm/day |

In the original WOFOST model, the measurements of soil, crop, agromanagement are assumed to be fixed with the time t while the measurement of weather is assumed to be variant. Thus, for the input vector P(t), we can divide it into two vectors as follows:

$$P(t) = \begin{bmatrix} P_1(t) \\ P_2 \end{bmatrix} \quad (2)$$

where the elements of $P_1(t)$ are the measurements of weather while the elements of $P_2$ are the measurements of soil, crop, agromanagement. To make the WOFOST model robust, we simulate the uncertainties of the measurements of soil, crop, agromanagement by creating M (M=50 in our research) different $P_2$ s by adding random Gaussian noise with distribution N(0,0.1) to it. For future use, let the added noise for each time be $\varepsilon_i (i \leq M)$ and we can obtain a noise vector E which length is M, which means this same perturbation E is applied to each time P(t). The E is denoted by:

$$E = \begin{bmatrix} \varepsilon_1 \\ \vdots \\ \varepsilon_M \end{bmatrix} \quad (3)$$

where $\varepsilon_i (i \leq M)$ is the i-th noise added to vector $P_2$. Thus, we obtain 50 new vectors P(t), denoted by

$$P(t, \varepsilon_i) = \begin{bmatrix} P_1(t) \\ P_2 + \varepsilon_i \end{bmatrix} \quad (4)$$

The prediction model with $P(t, \varepsilon_i)$ can be written as

$$L(\varepsilon_i, t+1) = WOFOST(L(0), L(\varepsilon_i, 1), ..., L(\varepsilon_i, t), P(\varepsilon_i, t)) \quad (5)$$

where L(0)=0, which is the same as the L(0) in the original model. The final prediction LAI value at t+1 could be obtained by

$$L(t+1) = \frac{\sum_{i=1}^{M} L(\varepsilon_i, t+1)}{M} \quad (6)$$

Experiments show the above model is better than the original model. However, accuracy is still an issue. Thus, we introduce data assimilation to improve the model.

*B. Data Assimilation for WOFOST model*

There are two major methods called EnKF and MCMC, which were often used for agriculture data assimilation. However, as we pointed out in the introduction, neither EnKF nor MCMC is effective when dealing with non-Gaussian distributions. However, activation functions in RNNS are usually nonlinear, such as tanh functions or ReLU functions. These nonlinear activation functions enable an RNN to handle nonlinear time series data better. Moreover, an RNN can be trained using backpropagation algorithms to optimize the parameters of the model by minimizing the loss function. The backpropagation algorithm enables an RNN to learn the mapping relationship from input to output, thus realizing the prediction of nonlinear time series data. Thus, we propose to combine recurrent neural network (RNN) and EnKF to solve the challenge and address the nonlinear issues typically encountered in traditional methods of agricultural data assimilation. We first will introduce data assimilation using EnKF, after that, we will introduce RNN.

1) **Data assimilation using EnKF**

For easy discussion, we define a vector $L_t = \begin{bmatrix} L(\varepsilon_1, t) \\ \vdots \\ L(\varepsilon_M, t) \end{bmatrix}$

which length is M to hold the predicted values at time t and matrix $A_t$ which is t*M to hold all of the predicted values from time 1 to time t as $A_t=[L_1, L_2, ..., L_t]$ (t=1,2...,N).

For EnKF, we also add noise to the observed LAI data. Let v(t) be the observed LAI value at time t. We create *M* noised v(t)s by adding $\varepsilon_i (i \leq M)$ to it and we obtain a vector $v_t$ from them as follows:

$$v_t = \begin{bmatrix} v(t) + \varepsilon_1 \\ \vdots \\ v(t) + \varepsilon_M \end{bmatrix} \quad (7)$$

We define a matrix $V_t$ which is t*M as follows:
$$V_t = [v_1, v_2, ..., v_t] \quad (t=1,2..., N) \quad (8)$$

For some dates of default observations, we set the $v_i$ corresponding to that date to $\begin{bmatrix} -1 \\ \vdots \\ -1 \end{bmatrix}$ to ensure that the matrix $V_t$ has the same dimension as the matrix $A_t$. The Kalman filter treats $v_j$ as a linear combination of the states represented by the operator *H*(t). There, *H*(t) represents the linear relationship between the simulation value and the observed value of LAI which can be expressed as

$$v_t = H(t)L_t + \sigma_t \quad (t=,1, 2, ..., N) \quad (9)$$

where $\sigma_t$ is a noise vector. *H*(t) is estimated from $v_t$ and $L_t$.

For data assimilation using EnKF, several matrices need to be computed. First, we need to estimate the covariance matrix *Re*(t), which can be calculated by using disturbance matrix *E* as follows

$$Re(t) = \frac{EE^T}{t-1} \quad (10)$$

where E can be obtained from eq (3), $E^T$ is the transpose of *E*.

Second, we need to estimate the variance matrix *Pe*(t), which will be used in Kalman filter. The computation is obtained by the matrix *A* as follows:

$$Pe(t) = \frac{A_t A_t^T}{t-1} \quad (11)$$

where $(A_t)^T$ is the transpose of $A_t$, and $\overline{A}_t$ is the conjugate matrix of $A_t$.

Using *Pe*(t), *Re*(t) and operator *H*(t), the Kalman gain *K*(t) can be calculated as followings:
$$K(t) = Pe(t)H(t)^T(H(t)Pe(t)H(t)^T + Re(t))^{-1} \quad (12)$$

For observation data V with stable values, linear operator H(t) can achieve data assimilation well and obtain optimal estimates. However, when the collected LAI data has a fluctuation value, H(t) may make the fluctuation more intense. *H*(t) is defined as the number of observed values that have not participated in assimilation at present. It is an identity matrix. *H*(t) is used to represent the linear relationship between the simulation state and the observed quantity.

The EnKF assimilated results can be calculated:
$$A_{t+1}^E = A_t + K(t)(V_t - H(t)A_t) \quad (13)$$

where the $A_{t+1}^E$ means the LAI result predicted by the EnKF method for day t.

In agricultural applications, because the observation network is sparse, the variability of the data is very significant in space, and the data is greatly affected by sampling error, the observation results are often far apart, resulting in inaccurate estimation of long-distance correlation or false correlation. At the same time, EnKF is prone to the potential underestimation of the covariance of prediction error when faced with sparse networks, which may lead to filtering bias and fail to effectively absorb new observations. Covariance localization and covariance inflation are therefore introduced to solve the problem of false correlation between distant points and the problem of underestimating the final uncertainty.

Covariance localization aims to eliminate the effect of spurious correlations among the state variables and the parameters by constraining the correlation range of the empirical covariance. This can be achieved by replacing Eq. (12) with the following equation:
$$K(t) = (\rho Pe(t)H(t)^T)(\rho(H(t)Pe(t)H(t)^T) + Re(t))^{-1} \quad (14)$$

where, ρ is determined using a fifth-order piecewise rational function, as given by[32]

$$\rho = \begin{cases} 1 - \frac{1}{4}(\frac{e}{l})^5 + \frac{1}{2}(\frac{e}{l})^4 + \frac{5}{8}(\frac{e}{l})^3 - \frac{5}{3}(\frac{e}{l})^2 & 0 \leq e \leq l \\ \frac{1}{12}(\frac{e}{l})^5 - \frac{1}{2}(\frac{e}{l})^4 + \frac{5}{8}(\frac{e}{l})^3 + \frac{5}{3}(\frac{e}{l})^2 - \frac{5}{3}(\frac{e}{l}) + 4 - \frac{2}{3}(\frac{e}{l})^{-1} & l \leq e \leq 2l \\ 0 & e \geq 2l \end{cases} \quad (15)$$

where e is the Euclidean distance between a grid point and an observation location and l is the horizontal influence radius.

Covariance inflation is a technique used to avoid filter divergence by inflating the empirical covariance. This can be achieved by linearly inflating each component of the augmented state vector *A*:

$$A_{i,t}^{inf} = \sqrt{\lambda_t}(A_{i,t} - \langle A_t \rangle) + \langle A_t \rangle \quad (16)$$

where $A_{i,t}^{inf}$ is the *i*-th ensemble member at the *t*-th time step of the LAI vector; $\langle \cdot \rangle$ denotes ensemble average; $\lambda_t$ is the inflation factor at the *t*-th time step. There are many methods to get the inflation factor $\lambda$, the method used in the work is the time-dependent inflation algorithm proposed by [33]:

$$\lambda_t = \frac{(Re(t)^{-\frac{1}{2}}d_t)^T Re(t)^{-\frac{1}{2}}d_t - k}{\text{trace}\{Re(t)^{-\frac{1}{2}}H(t)Pe(t)(Re(t)^{-\frac{1}{2}}H(t))^T\}} \quad (17)$$

where k is the number of observations; $d_t$ is the residual between observation data and forecast data, which can be described as:

$$d_t = v_t - H(t)\langle A_t \rangle \quad (18)$$

Finally, the EnKF assimilated results $A_{t+1}^E$ can be calculated by combining the covariance localization and the covariance inflation:

$$A_{t+1}^E = A_t^{inf} + K(t)(V_t - H(t)A_t) \quad (19)$$

where $A_t^{inf}$ is the matrix which contains every $A_{i,t}^{inf}$, the $A_{t+1}^E$ means the LAI result predicted by the EnKF method for day t. $A_{t+1}^E$ has the same size as $A_t$, and the assimilated result of $L_t$ is the t-column of $A_{t+1}^E$. Let the t-column of $A_{t+1}^E$ be $L_{t+1}^E$ and thus $L_{t+1}^E$ is the assimilated result of $L_t$. When estimating $L_{t+2}$, the assimilated result $L_{t+1}^E$ will be used to replaced $L_{t+1}$ as the input of the WOFOST model. Figure 2 shows the flowchart for LAI prediction with EnKF assimilation. LAI prediction with EnKF assimilation can obtain better accuracy, However, there are two limitations on EnKF methods: (1) EnKF method can only be effective within the time interval where the data in V are available. If there is too much fault data in V at some time interval, the method has less effect on data assimilation of LAI; (2) EnKF process still uses linear operators H, causing problems in assimilating nonlinear data. Therefore, the second stage of assimilation, LSTM, is carried out.

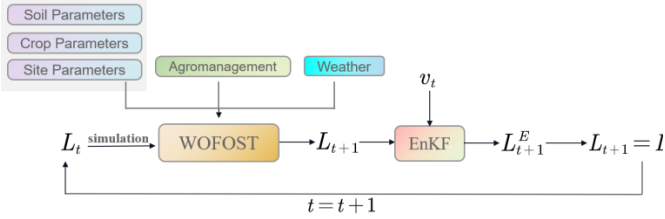

**Fig. 2. (a).** The EnKF Part of the Proposed Prediction Model

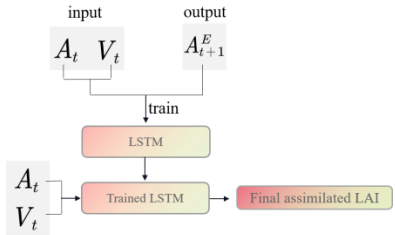

**Fig. 2. (b).** The LSTM Part of the Proposed Prediction Model

2) **Data Assimilation with EnKF-LSTM**

In the first stage, the EnKF method assimilates and predicts the simulated data generated by WOFOST. The formula (19) can be viewed as A function of A and V:

$$A_i^E = f(A_{i-1}, V_{i-1}) \quad (20)$$

where the $A_i^E$ means the LAI result predicted by the EnKF method for day i. The neural network can capture the weights and mappings hidden in the ensemble Kalman filtering method, so in the second stage, LSTM will be used to learn the assimilation method of ensemble Kalman filtering to achieve the enhancement of EnKF assimilation results. For each time T we need a new network to perform the prediction at time t+1 so we train a neural network at each time t. During the training of the neural network, each training data consists of three parts: the simulation vector $L_i$ ($0 < i \leq t$) generated by WOFOST, the observation vector $v_i$, and the assimilated data after EnKF $L_i^E$. We hope that the LSTM neural network can obtain results close to EnKF by input $L_i$ and $v_i$. For this purpose, we designed a neural network with the features of Figure 3. The input of the neural network is divided into two parts: the WOFOST prediction matrix A and the observation matrix V. Each day's prediction vector $L_i$ and observation vector $v_i$ ($0 < i \leq t$) form a matrix $(L_i, v_i)$, and each $(L_i, v_i)$ corresponds to an input neuron, forming an input layer. The number of output neurons is the number of variables to assimilate, i.e., t. The input $(L_i, v_i)$ of day i will get the prediction result of day i+1, which is $L_{i+1}$, after the neural network operation. Output neurons form the output layer.

The neural network also contains multiple hidden layers between the input layer and the output layer. Network training needs to set the number of hidden layers, the number of neurons, the activation function, and the learning rate parameters. Where parameter $h_i^{(k)}$ represents the output value of the current unit, $c_i^{(k)}$ represents the neuronal memory of the current unit, i indicates that the parameter is the output and memory state generated by the input $(L_i, v_i)$, k indicates that the parameter is on the k-th hidden layer. The parameters $h_i^{(k)}$ and $c_i^{(k)}$ of each layer of the neural network are initialized to the $\vec{0}$.

All outputs $L_i^D$ of the neural network form matrix $A_i^D$,

$$A_i^D = \sigma(W_i(\dots\sigma(W_1[A_i, V_i] + b_1)) + b_i)$$
$$= g(A_i, V_i) \quad (21)$$

where $W_i$ represents the weight vector of the output gate on day i, and $b_i$ represents the bias vector of the output gate on day i.

Because of $A_i^D \approx A_i^E = f(A_i, V_i)$, so $g \approx f$.

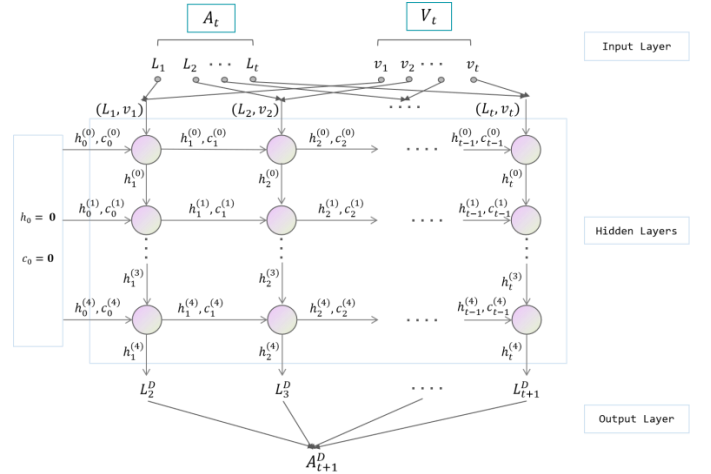

**Fig. 3.** The structure of LSTM neural network.

Loss function represents the Euclidean Distance between the current model's output vector and the target vector. The back-propagation algorithm by the gradient descent criterion is used to reduce the loss.

This method of using the data results of EnKF data assimilation to train the LSTM neural network effectively reduces the model error generated in each time step $\triangle t$.

The solution of EnKF-LSTM algorithm is also more accurate than the solution of EnKF algorithm. Let $\delta_k^E$ denote the error in EnKF algorithm, $\delta_k^D$ denote the error in EnKF-LSTM algorithm, it has been proved by [26] that:

$$\|\delta_k^D\|_\infty \leq \|\delta_k^E\|_\infty \quad (22)$$

Based on the learning law of EnKF data assimilation, LSTM can assimilate simulated data A well on the default date of observed data V, which avoids the problem of EnKF exacerbating the fluctuation of observed data. Especially when processing the nonlinear data near the inflection point of LAI growth curve, the neural network can more accurately find the position where the maximum value should be and solve the fluctuation in EnKF assimilation. The background error of EnkF-LSTM algorithm is smaller than that of EnKF algorithm, and the actual results obtained are more accurate. Compared with EnKF method, it can be proved that the data assimilation method with superimposed LSTM neural network has better performance.

IV. EXPERIMENTS

*A. Experimental site*

To ensure data quality, LAI should be collected on cloudy days or before sunrise or after sunset. On some days, the weather such as sun exposure and rainstorm are not suitable for LAI measurement. Then, some values of LAIs are default at some special days. To make the observed value align to the simulated value at each date, some null placeholders are used on a date when there is no value of LAI.

The experimental field is set at a farm located in Heilongjiang province, northern China, covering a total area of 53,100 square kilometers. A total of 12 groups of sensor devices were deployed in the experimental field. Each group of sensor devices uploads the environmental variables every half hour to the data center. Each record contains 16 columns, including 14 features which are concentrations of $CO_2$, soil temperature and soil relative humidity at the surface and below 10, 20, 30 and 40 cm, air temperature, air humidity, light intensity. In addition to that, there are collection time, and sensor device number. LAI of crops was manually collected at 2–7-day intervals between June 18 and September 15.

*B. Experimental design*

The data was collected on the farm from April 2022 to September 2022. The environmental data include carbon dioxide index, soil temperature (including surface layer, 20cm, 30cm, 40cm, 50cm), soil moisture (including surface layer, 20cm, 30cm, 40cm, 50cm), relative humidity, etc. The crop growth data is leaf area index (LAI).

Python Crop Simulation Environment (PCSE), a python implementation of WOFOST, was used. As mentioned above, a default file containing parameters for maize, rice and soybean was selected, and open-source weather data was used to generate simulated crop data as our source domain dataset. Specifically, the PCSE simulated crop growth cycle. The virtual maize, rice and soybean crop was set to be sowed on May 1st and harvested on October 15th, 2022. As a result, the total time of simulated growth was 168 days. We generated over 150 rounds of the complete growth process of virtual maize, rice, and soybean. The output of the PCSE had 25200 rows in total and was considered as time-series data where each element in the series was a multi-dimensional vector. The vector represents all the features concerning weather conditions and the crop itself, including Leaf Area Index (LAI) which characterizes the plant growth situation. It is defined as the green leaf area per unit ground surface area, and it served as an essential feature for crop growth prediction.

*C. Validation methods*

MSE (Mean Square Error), RMSE(Root Mean Square Error), MAE(Mean Absolute Error) are used as predictive performance metrics.

1) **MSE**

MSE is the difference between the real value and the predicted value squared and then averaged. It is often used as a loss function of linear regression because it is easy to differentiate by the square form. The formula of the MSE is:

$$\text{MSE} = \frac{1}{N} \sum_{i=1}^{n}(Y_i - f(x_i))^2 \qquad (23)$$

The statistical parameter is the meaning of the sum of squares of the errors of the points corresponding to the predicted data and the original data. Where Yi represents the true value and f(xi) represents the predicted value.

2) **RMSE**

RMSE stands for root mean square error, which is the square root of the ratio of the squared deviation between the predicted value and the actual value and the number of observations n, the calculation formula is:

$$\text{RMSE} = \sqrt{(\frac{1}{N} \sum_{i=1}^{n}(Y_i - f(x_i))^2)} \qquad (24)$$

RMSE measures the deviation between the predicted and the actual value. It is sensitive to outliers in the data. When the number of times is higher, the more relevant the calculation is to the larger values. Then the smaller values are ignored.

3) **MAE**

MAE is the average absolute error between the predicted value and the observed value. The calculation formula is:

$$\text{MAE} = \frac{1}{N} \sum_{i=1}^{n}|Y_i - f(x_i)| \qquad (25)$$

MAE is a linear score that is averaged directly over residuals and all individual differences are weighted equally over the mean.

*D. Parameters Setting*

To evaluate the effect of ENKF-LSTM data assimilation method, this paper compares its results with the initial simulation results, EnKF assimilation results, and the data assimilation results of CNN, GRU and FNN neural networks. The parameter Settings of different neural networks are shown in the following table:

TABLE II
PARAMETERS OF DIFFERENT NNS

| Neural Network | Parameters | Values |
|---|---|---|
| GRU | C | 3.0 |
| | Epsilon | 0.000001 |
| | Degree | 3 |
| | Tolerance | 0.00001 |
| CNN | Total Layers | 8 |
| | Rate of Learning | 0.001 |
| | Optimization Approach | Adam |
| | Loss | binary_crossentropy |
| | Epochs | 100 |
| | Batch Size | 64 |
| | Activation | Relu |
| FNN | Total Layers | 3 |
| | Rate of Learning | 0.001 |
| | Optimization Approach | Adam |
| | Batch Size | 32 |
| | Epochs | 200 |
| LSTM | Total Layers | 3 |
| | Neurons | (16,32,64,128) |
| | Rate of Learning | 0.001 |
| | Optimization Approach | Adam |
| | Size of the Batch | 10 |
| | Epochs | 300 |
| | Step down time | 3 |

*E. Results and Discussion*

The above data assimilation methods respectively conducted data assimilation on the 2022 growth data of rice, corn, and soybean. The assimilation results are shown in the figure below. The performance pairs of different methods are shown in the table.

TABLE III
RESULTS OF RICE EXPERIMENTS ON DIFFERENT METHODS

| | | MSE | RMSE | MAE |
|---|---|---|---|---|
| Rice | Initial | 13.58284178 | 3.685490711 | 3.474919768 |
| | EnKF | 3.815672926 | 1.953374753 | 1.699443445 |
| | EnKF-LSTM | 0.446925175 | 0.668524626 | 0.411223975 |
| | CNN | 0.499665677 | 0.70687034 | 0.5097246 |
| | GRU | 1.497539573 | 1.223739994 | 1.012123917 |
| | FNN | 2.946442552 | 1.716520478 | 1.457400705 |

For the results of rice crops in table 3, the assimilation results of the EnKF-LSTM method, compared to the initial data, showed a decrease of 96.71% in MSE, 81.86% in RMSE and 88.17% in MAE. Compared to the EnKF method without using neural networks, the MSE decreased by 88.29% and the RMSE decreased by 65.78%, the MAE decreased by 75.80%. This is because EnKF relies on prior models and observational models and has limitations when dealing with nonlinear problems. On the other hand, LSTM is capable of handling complex nonlinear relationships, allowing for better handling and imputation of noise and missing data.

The EnKF-LSTM method showed a decrease of 10.56% in MSE, 5.42% in RMSE, and 19.32% in MAE compared to CNN, and a decrease of 84.83% in MSE, 61.05% in RMSE, and 71.78% in MAE compared to FNN. For FNN, which is a feedforward neural network with no direct connection between the input and output of each time step, sequence features cannot be captured effectively when processing data, which makes accurate prediction difficult. Compared to the GRU neural network, the EnKF-LSTM method exhibited a decrease of 70.16% in MSE, 45.37% in RMSE, and 59.37% in MAE. This is because LSTM's gating mechanism is more complex, making it better suited for handling intricate sequential patterns and long-term dependencies.

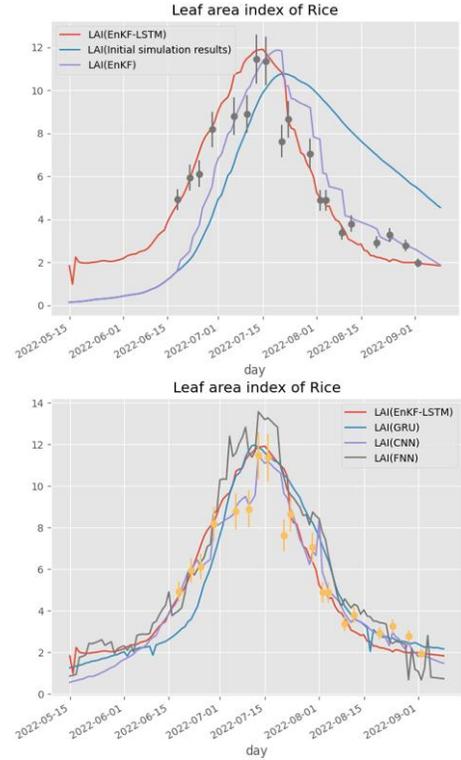

**Fig. 4.** LAI of Rice under different models.

TABLE IV
RESULTS OF MAIZE EXPERIMENTS ON DIFFERENT METHODS

| | | MSE | RMSE | MAE |
|---|---|---|---|---|
| Maize | Initial | 0.414495257 | 0.64381306 | 0.601347162 |
| | EnKF | 0.250176871 | 0.50017684 | 0.425573537 |
| | EnKF-LSTM | 0.031018421 | 0.176120474 | 0.145141995 |
| | CNN | 0.17682105 | 0.420500951 | 0.284707482 |
| | GRU | 0.150129302 | 0.387465226 | 0.351227713 |
| | FNN | 0.178231412 | 0.422174623 | 0.362140680 |

In the experiment with Maize crops, table 3 shows that the EnKF-LSTM method exhibited a decrease of 92.52% in RMSE, 72.64% in MSE, and 75.86% in MAE compared to the unassimilated data. Compared to the EnKF method, the EnKF-LSTM method showed a decrease of 87.60% in RMSE, 84.79% in MSE, and 65.89% in MAE according to the data in table 3. Compared to CNN, the EnKF-LSTM method achieved decreases of 82.46%, 58.11%, and 49.02% in MSE, RMSE, and MAE. Compared to GRU, the EnKF-LSTM method achieved decreases of 79.34%, 54.55%, and 58.68% in MSE, RMSE, and MAE. Compared to FNN, the EnKF-LSTM method achieved decreases of 82.60%, 58.28%, and 59.92% in MSE, RMSE, and MAE. This is because the observed data for maize crops are only available in the middle of growth period from June 18 to July 22, 2022. This makes it more challenging for CNN, GRU and FNN to extract temporal information and handle long-term dependencies. Therefore, LSTM's advantages are more prominent when dealing with maize data.

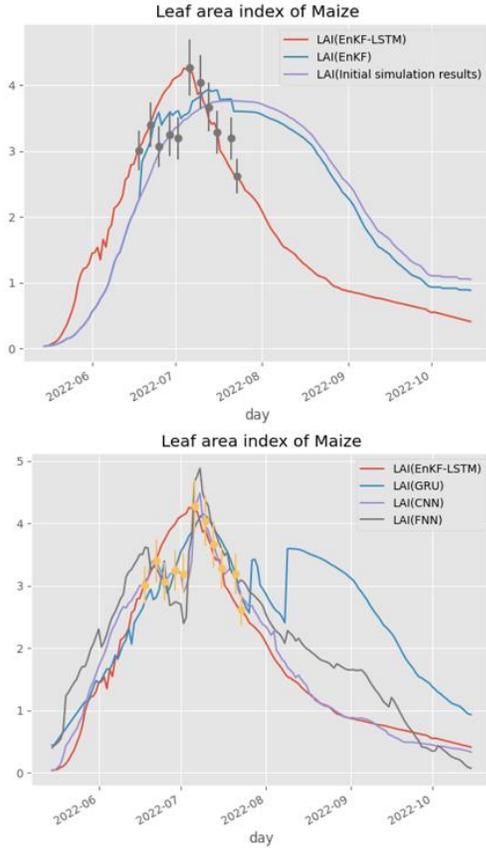

**Fig. 5.** LAI of Maize under different models.

TABLE V
RESULTS OF SOYBEAN EXPERIMENTS ON DIFFERENT METHODS

|  |  | MSE | RMSE | MAE |
|---|---|---|---|---|
|  | Initial | 0.300224984 | 0.5479279 | 0.433973048 |
|  | EnKF | 0.15597515 | 0.394936894 | 0.300883227 |
| Soybean | EnKF-LSTM | 0.047868051 | 0.218787685 | 0.173885064 |
|  | CNN | 0.094854378 | 0.307984379 | 0.207569351 |
|  | GRU | 0.157369207 | 0.396697878 | 0.331384619 |
|  | FNN | 0.2073425927 | 0.455348870 | 0.3164686439 |

Similarly, when dealing with soybean crops, the EnKF-LSTM method maintains a significant advantage. According to the results in table 3, compared to EnKF, the EnKF-LSTM method achieved decreases of 93.75% in MSE, 74.99% in RMSE, and 42.21% in MAE. The observed values of soybean are relatively evenly distributed, but there are still some abnormal values (such as on July 15) that are difficult for EnKF to handle. MSE and RMSE are significantly reduced because of the robustness of LSTM neural networks in processing noise. Compared to CNN, GRU and FNN neural networks in Figure 6(b), The EnKF-LSTM method results better because the LSTM network has a complex gated unit and can handle long-term dependencies.

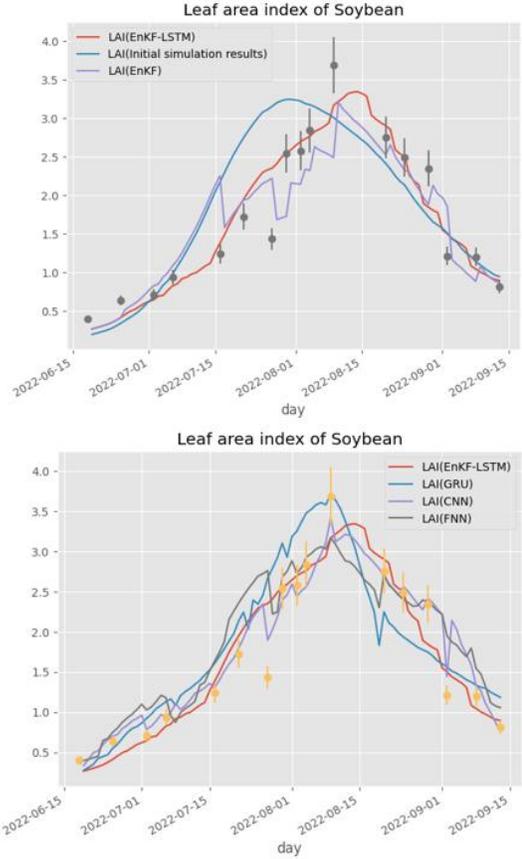

**Fig. 4.** LAI of Soybean under different models.

TABLE VI
COMPARISON OF DIFFERENT CROPS BY ENKF-LSTM

| Crop | MSE | RMSE | MAE |
|---|---|---|---|
| Rice | 0.446925175 | 0.668524626 | 0.411223975 |
| Maize | 0.031018421 | 0.176120474 | 0.145141995 |
| Soybean | 0.047868051 | 0.218787685 | 0.173885064 |

The above three experiments reflect three different observational data styles. For the uniform and accurate observation data of rice crops, the difference between different data assimilation methods is not significant. The observation data of maize crops are only available at the middle of growth period, and EnKF-LSTM method has obvious advantages in comparative experiments because of its ability to deal with long-term dependence. For soybean data with uniform distribution but large error, EnKF-LSTM method also has advantages in processing soybean data because of its robustness to noise.

V. CONCLUSIONS AND FUTURE WORK

In this study, a ENKF-LSTM data assimilation method is proposed for WOFOST crop growth model. The LAI growth curve prediction was realized for three crops of rice, maize, and soybean. Experiments show that the LAI prediction method has a significant effect, which can effectively fuse the WOFOST prediction results with the measured data of sensors. Compared with the method of EnKF directly acting on the WOFOST crop growth model, there is no obvious overfitting phenomenon. The loss function value is significantly reduced, which effectively solves the problems existing in the traditional method. The assimilation effect is considerable. Although the method design and three experiments were carried out using LAI data as an indicator, this method also has guiding significance for the measurement of other crop growth, such as crop biomass and leaf dry weight. They can be predicted and corrected by the method introduced in this paper. This method has a wide range of applications in the agricultural field.

ACKNOWLEDGMENT

This research is supported by China New Generation Artificial Intelligent Program(No.21ZD0110900), Heilongjiang.

NSF funding in China (No.LH202F022), and the Fundamental Research Funds for the Central Universities in China (No. 2023FRFK06013).